\documentclass[runningheads]{llncs}
\usepackage{graphicx}
\usepackage{comment}
\usepackage{booktabs}

\usepackage[utf8x]{inputenc}
\usepackage[ruled,vlined,linesnumbered]{algorithm2e}
\usepackage{amsmath,amssymb} 
\usepackage{color}

\usepackage{blindtext}
\usepackage{floatrow}
\newfloatcommand{capbtabbox}{table}[][\FBwidth]

\newcommand{\f}{\textbf{\textit{f}}}

\begin{document}
\pagestyle{headings}
\mainmatter
\def\ECCVSubNumber{4042}  

\title{Learning from Scale-Invariant Examples for Domain Adaptation in Semantic Segmentation} 

\titlerunning{Learning from Scale-Invariant Examples}
%
\author{M. Naseer Subhani \and
Mohsen Ali }
\authorrunning{M.Naseer Subhani \and
Mohsen Ali}
%
\institute{Information Technology University, Pakistan \\
\email{\{msee16021,mohsen.ali\}@itu.edu.pk}}

\maketitle

\begin{abstract}
Self-supervised learning approaches for unsupervised domain adaptation (UDA) of semantic segmentation models suffer from challenges of predicting and selecting reasonable good quality pseudo labels. 
In this paper, we propose a novel approach of exploiting  \textit{scale-invariance property} of the semantic segmentation model for self-supervised domain adaptation. 
Our algorithm is based on a reasonable assumption that, in general, regardless of the size of the object and stuff (given context) the semantic labeling should be unchanged. 
We show that this constraint is violated over the images of the target domain, and hence could be used to transfer labels in-between differently scaled patches. 
Specifically, we show that semantic segmentation model produces output with high entropy when presented with scaled-up patches of target domain, in comparison to when presented original size images. These scale-invariant examples are extracted from the most confident images of the target domain. Dynamic class specific entropy thresholding mechanism is presented to filter out unreliable pseudo-labels. Furthermore, we also incorporate the focal loss to tackle the problem of class imbalance in self-supervised learning.
Extensive experiments have been performed, and results indicate that exploiting the scale-invariant labeling, we outperform existing self-supervised based state-of-the-art domain adaptation methods. Specifically, we achieve 1.3\% and 3.8\% of lead for GTA5 to Cityscapes and SYNTHIA to Cityscapes with VGG16-FCN8 baseline network.


\end{abstract}

\section{Introduction}

Deep learning based semantic segmentation models \cite{yu2017dilated,chen2018deeplab,zhao2017pyramid,zhao2018icnet} have made considerable progress in last few years. 
Exploiting hierarchical representation, these models report state-of-the-art results over the large datasets. 
However, these models do not generalize well; when presented with out of domain images, their accuracies drops. 
This behavior is attributed to the shift between the source domain, one over which model has been trained, and target, over which its being tested. 
Most of semantic segmentation algorithms are trained in a supervised fashion, requiring pixel-level, labor extensive and costly annotations. Collecting such fine-grain annotations for every scene variation is not feasible.
To avoid this pain-sticking task, road scene segmentation algorithm use synthetic but photo-realistic datasets, like GTA5 \cite{richter2016playing}, Synthia  \cite{ros2016synthia}, etc., for training. However, they are evaluated on the  real datasets like Cityscapes \cite{cordts2016cityscapes}, thus amplifying the domain shift.

\begin{figure}[t]
\begin{center}
\includegraphics[width=1\linewidth]{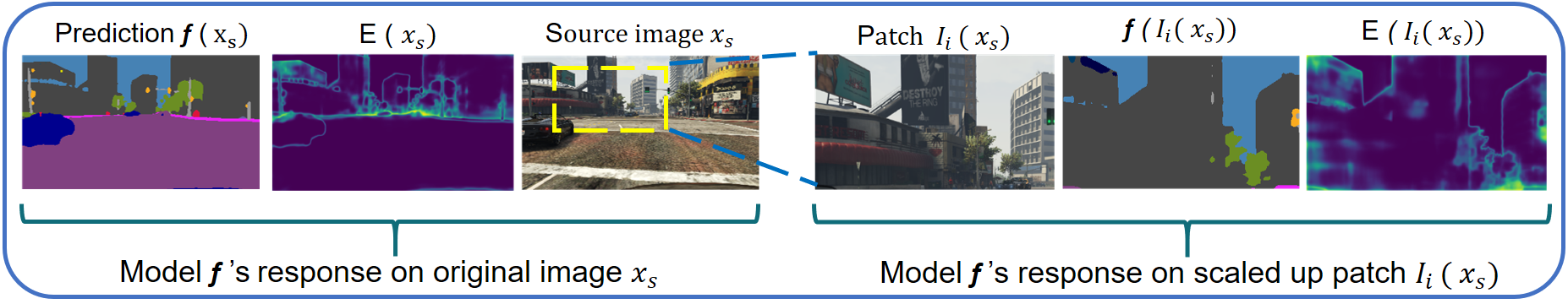}
\end{center}
  \caption{ \textbf{Scale-invraince property of semantic segmentation model} Original image and patch extracted from it and resized, are assigned same semantic labels by the model $\f$ at the corresponding locations. \textit{Left:} An image $x_s$ from the source domain, labels assigned to it by model $\f$. $x_s$ belongs to the source domain. Self-entropy map $E$ shows small values. Yellow box on $x_s$ indicate patch location \textit{Right:} Extracted patch resized to original image size. Assigned labels are similar to ones of original and self-entropy is similar that of original image. 
  }
\label{fig:scale_inv_source}
\end{figure}

Over the years, many unsupervised domain adaptation (UDA) methods have been proposed to overcome the domain shift, employing adversarial learning \cite{chen2017no,ganin2016domain,sankar2017unsuper,zhu2017unpaired}, self-supervised learning \cite{zhao2017pyramid,zou2018unsupervised,Iqbal_2020_WACV}, etc. or their combination. Where adversarial learning methods are dependent upon how good (input, feature or output) translation could be performed, self-supervised learning methods have to deal with challenges of generating so-called \textit{good quality} pseudo-labels and selection of confident images for the learning from the domain.

In this paper we propose a novel method of generating pseudo-labels for self-supervised adaptation for semantic segmentation, by exploiting scale-invariance property of the model.  Our proposed solution is based on an assumption that regardless of the size of an object in the image, the model's prediction should not be change, as shown in Fig. \ref{fig:scale_inv_source}. 
To support our algorithm, we introduce three other novel components to be incorporated in the self-supervised method.
A \textit{class-based sorting mechanism} image selection process to identify images that should be used for the self-learning. 
To filter out pixels with non-confident pseudo-labels from learning process, we design an automatic process of estimating \textit{class specific dynamic entropy-threshold} allowing "easy" classes to have tighter threshold than the ones that are ''difficult" to adapt.
To further reduce the effect of class imbalance over adaptation process, we also incorporate the focal loss \cite{lin2017focal} in our loss. Below we define the concept of scale-invariance.  

\begin{figure}[t]
\begin{center}
\includegraphics[width=1\linewidth]{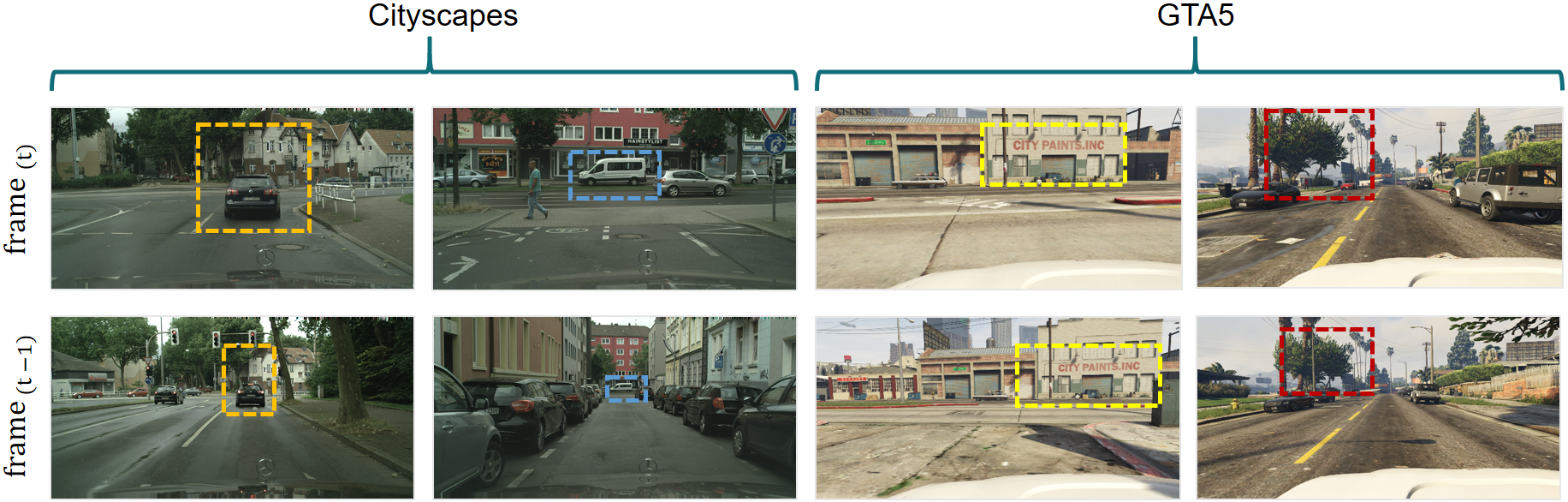}
\end{center}
  \caption{Objects and scene-elements exhibit the scale variations naturally in road scene images, as shown in the frames sampled from Cityscapes \cite{cordts2016cityscapes} and GTA5\cite{richter2016playing} datasets. As the vehicle moves, near by objects and other scene elements might become afar or vice-versa, resulting in scale changes. Matching color boxes highlight changing size of cars, buildings, and other regions as vehicle moves.  
  }
\label{fig:scale_inv_effect}
\end{figure}

\textit{Scale-invariance: } In general one can assume that depending on the camera location, pose and other parameters, objects in images will appear at varying sizes.
In the road scene imagery, such as GTA5, Cityscapes, etc., due to movement of the vehicle and dynamic nature of environment, objects and other scene elements (like road, building) appear at multiple scales.  
These variations are readily visible in Fig. \ref{fig:scale_inv_effect}.
Its reasonable to assume that the semantic segmentation model trained on such dataset that will assign objects and stuff with same semantic labels regardless of their size.
This could be seen in Fig. \ref{fig:scale_inv_source}, where when an image and a resized patch extracted from same image are presented to segmentation model we get similar  semantic labels at (almost all) corresponding regions. 
For both, image and resized patch, self-entropy is also indicating that the decision was made with low uncertainty. 
Semantic segmentation model, when presented with an image, from the out of source but somewhat visually similar domain, and the patches extracted from that image, we see considerable difference between the labels assigned for patches and ones assigned to corresponding areas of original image. 
Comparative increase in the self-entropy indicates that labels assigned to patches are not reliable . 
In this work, as shown in Fig. \ref{fig:targ_sci}, we propose to use semantic labels assigned to the image to create pseudo-labels of corresponding patches. Our objective is to preserve the scale-invariance property of the semantic segmentation model and use it to direct our adaptation process. \\


We summarize our contribution as bellow. 
\begin{itemize} 
\item  We propose a novel approach of exploiting scale-invarince property of the model to generate pseudo-labels for the self-supervised domain adaptation of semantic segmentation model.
\item Class specific dynamic entropy thresholding is introduced so that pixels belonging to classes at different adaptation stage could be judged differently when being made included in the loss function. 
\item To eliminate the effect of the class imbalance problem, we incorporate the focal loss to boost the performance of smaller classes. And Class-based target image sorting algorithm is proposed so that selected images have equal representation of all the classes. 
\end{itemize}

Although, part of our algorithm is generic, we show our results on the adaptation from synthetic to real road scene segmentation. We report state-of-the-art results over the GTA to Cityscapes and Synthia to Cityscapes for the self-supervised based domain adaptation algorithms. VGG16 \cite{simonyan2014very} and ResNet101 \cite{he2016deep} are used as our baseline architectures.

\section{Related works}
\textbf{Semantic Segmentation:} There is an intensive amount of research has been done in semantic segmentation due to its importance in the field of computer vision. State of the art methods in semantic segmentation have gained huge success for their contribution. Recently, many researchers have proposed algorithm for semantic segmentation such as DRN (Dilated Residual Network) \cite{yu2017dilated}, DeepLab \cite{chen2018deeplab} etc. \cite{badrinarayanan2017segnet,zhao2017pyramid,wang2018understanding}. \cite{yu2017dilated} have proposed a dilated convolution neural network in semantic segmentation to increase the depth resolution of the model without effecting its receptive field. In this work, we have utilized FCN8s\cite{long2015fully} with VGG16\cite{simonyan2014very} and DeepLab \cite{chen2017deeplab} with ResNet101 \cite{he2016deep} as our baseline architectures of semantic segmentation.\\\\
\textbf{Domain Adaptation:} Domain adaptation is a popular research area in computer vision, especially in classification and detection problems. The goal of domain adaptation is to minimize the distribution gap between source and target domain. Many of the algorithms have already developed for domain adaptation like \cite{zou2018unsupervised,vu2018advent,sankaranarayanan2018learning,hoffman2017cycada,tsai2018learning,zhang2017curriculum,hoffman2016fcns,zhu2017unpaired,Iqbal_2020_WACV}. In this paper, we are focused in self-supervised domain adaptation to tackle the problem of domain diversity. Previous methods have been applied Maximum Mean Discrepancy (MMA) \cite{quinonero2008covariate} to minimize the distribution difference. Recently, there has been an enormous interest in developing domain adaptation methods with the help of unsupervised and semi-supervised learning.\\\\
\textbf{Adversarial Domain Adaptation in Semantic Segmentation:}
Adversarial training for unsupervised domain adaptation is the most explored approach for semantic segmentation. \cite{hoffman2016fcns} are the first ones to introduce domain adaptation in semantic segmentation. \cite{vu2018advent} have proposed an entropy minimization, based on domain adaptation in which they have minimized the self-entropy with the help of adversarial learning. In \cite{tsai2018learning}, they have applied adversarial learning at the output space to minimize the distribution at the pixel level between the source and the target domain. \cite{chen2018road} presents Reality-Oriented-Adaptation-Network (ROAD) to learn invariant features of source and target domain by target guided distillation and spatial-aware adaptation. \cite{luo2019taking} has also introduced a categorical-level adversarial network (CLAN) in which they have aligned the features of each class by adaptive adjusting the weight on adversarial loss specific to each class. There are other methods with the generative part for adversarial training in semantic segmentation.  In generative methods, they are trying to generate the target images with a condition of the source domain. \cite{zhu2017unpaired} have proposed a pixel level adaptation to generate image similar in visual perception with target distribution. In \cite{hoffman2017cycada}, they have used pixel level and feature level adaptation to overcome the distribution gap between the source and the target domain. They incorporate cycle consistency loss to generate the target image condition on the source domain. They have also utilized the feature space adaptation and generate target images from the source features and vice-versa.\\\\
\textbf{Self-Supervised Domain Adaptation in Semantic Segmentation:}
The idea behind self-supervised learning is to adapt the model by the pseudo labels generated for unlabeled data from the previous state of the model. 
\cite{laine2016temporal} proposed a method of self-supervised learning from the assembling of the output from different models and latter train the model by generating pseudo labels of unlabeled data. \cite{tarvainen2017mean} developed an algorithm based on a teacher network where the model is adapted by averaging the different weights for better performance on the target domain. Recently, self-supervised learning has also gained popularity in the semantic segmentation task. \cite{zou2018unsupervised} proposed a class-balanced-self-training (CBST) for domain adaptation by generating class-balanced pseudo-labels from images which were assigned labels with most confidence by last state of model. 
To help guide the adaptation, spatial priors were incorporated. 
 \cite{du2019ssf}  have also contributed their research in self-supervised learning by generating pseudo labels with a progressive reliable strategy. They have excluded less confident classes with a constant threshold and have trained the model on generated pseudo labels. In this research, we filter out the less confident classes by applying a dynamic threshold that is calculated for each class separately during the training process. \cite{lian2019constructing} have proposed a self-motivated pyramid curriculum domain adaptation (PyCDA) for semantic segmentation. They have included the curriculum domain adaptation by constructing the pyramid of pixel squares at different sizes, which has included the image itself. The model trained on these pyramids of the pixel by capturing local information at different scales. Iqbal and Ali \cite{Iqbal_2020_WACV}'s  spatially independent and semantically consistent (SISC) pseudo-generation method could be closest to our work. However, they only explore the spatial invariance by creating multiple translated versions of same image. 
Since they don't have knowledge of which version has results in better inference they aggregate inference probabilities from all to create a single version, leading to smoothed out pseudo-labels. 
We on the other hand, define a relationship between the scale of the image and the self-entropy; therefore instead of aggregating we use the inference for image of original scale to create pseudo-labels for the up-scaled patch extracted from same image. 
Along with it, we present a comprehensive strategy of overcoming class imbalance and selecting the reliable psuedo-labels.

\section{Methodology}
In this section, we briefly describe our propose domain adaptation algorithm by learning from self-generated scale-invariant examples for semantic segmentation.
In this work, we assume that the predictions of these confident images on target data are the approximation of their actual labels.
\subsection{Preliminaries}
Let $X_S$ be set of images belonging to the source domain, such that for each image  $x_s \in {\mathbb{R}}^{H \times W \times 3}$, in the source domain we have respective ground-truth one-hot encoded matrix  $y_s \in {\mathbb{R}}^{H \times W \times C}$. Where $C$ is the number of classes and $H \times W$ is the spatial size of the image.
Similarly, let $X_T$ be set of images belonging target domain.
We train a fully convolution neural network, $\f$, in a supervised fashion over the source domain for the task of semantic segmentation. 
Let $P=\f(x)$ be soft-max output volume of size $H \times W \times C$, representing predicted semantic class probabilities for each pixel. The segmentation loss for any image $x$ with the given ground-truth labels $y$ and predicted probabilities $P$ is given by 
\begin{equation}
\mathcal{L}_{seg}(x, y) = - \sum_{h,w,c}^{H, W, C} y^{h,w,c} log(P^{h,w,c})
\label{eq_a_}
\end{equation}
In later cases to increase readability we just use $h,w,c$ with summation sign, to indicate the summation over total height, width and channels.
Source model $\f$ has been trained by minimizing $\mathcal{L}_{seg}^S=\sum_{s}^{S}\mathcal{L}_{seg}(x_s, y_s)$.

For target domain, since we do not have ground-truth labels, self-supervised learning method requires us to generate \textit{pseudo-labels}. 
Let $x_t \in X_T$ be an image in the target domain, $P_t=\f(x_t)$ be output probability volume, one hot encoded pseudo-labels $\hat{y}_t$ could be generated by assigning label at each pixel to the class with maximum predicted probability. 
Since, source model is not accurate on the target domain, therefore a binary map $F_t \in {\mathbb{B}}^{H \times W}$ is defined to select the pixels whose prediction loss has to be back-propagated. 
\begin{equation}
\mathcal{L}_{seg}(x_t, \hat{y}_t) = - \sum_{h,w,c}^{H, W, C} F_t^{h,w} \hat{y}_t^{h,w,c} log(P_t^{h,w,c})
\label{eq_selfloss}
\end{equation}
In general, for self-supervised learning, we minimize the loss in Eq. \ref{eq_selfloss} over the selected subset of images from the target domain. 


\subsection{Class-Based Sorting for Target Subset Selection}
\label{sec:classSort}
To train the model with self-supervised learning, we need to extract the pseudo-labels which are reliable. 
A binary filter defined in Eq. \ref{eq_selfloss}, helps select pixels with so-called \'good\'    pseudo-labels, however, does not give us global view of how good are predictions in the whole image. 
Calculating an average of maximum probability per location of $\hat{y}_t$ can help us define the confidence of predictions on the $x_t$, for readability we call it \textit{confidence of image $x_t$}. A subset selected on the base of the above defined confidence can lead to a class-imbalance with more images with pseudo-labels belonging to large and frequently appearing classes. That in turn leads to adaptation failing for the smaller objects or infrequent classes. 
We design a class based image subset selection process from the target domain (Algo. \ref{image_selection_alg}) to mitigate this effect. 

Instead of calculating confidence for each image globally, we calculate the \textit{confidence} with respect to each class \textit{c} for every image in target data \textit{X}\textsubscript{T}. 
For each class, $X_T$ is sorted with respect to the class specific confidence $U_c$ and a subset, of size $p$, is selected.
Union of these subsets form our confident target images subset $X_{t}^{'}$, note that repeated entries are removed. The algorithm of class-based sorting shown in Algorithm \ref{image_selection_alg}.
For $X^{'}_t$ the model prediction are relatively of more confidence than rest of the set and can be utilized to adapt the model by self-supervised learning. 

\begin{algorithm}[t!]
\DontPrintSemicolon 
\SetAlgoLined
\SetKwInOut{Input}{Input}\SetKwInOut{Output}{Output}
\Input{Model \textit{f}(\textbf{w}), Target data \textit{X}\textsubscript{t},  portion \textit{p}}
\Output{Confident images {\textit{X$'$}}\textsubscript{t} of target domain , Entropy threshold \textit{h}\textsubscript{c} }
\BlankLine
\For{t = 1 to T}{
   $P_{x_{t}} = \textbf{\textit{f}}\,(w,x_{t})$\;
    $M_{P_{x_{t}}} = max\,(P_{x_{t}},\,axis =  0)$\;
    $A_{P_{x_{t}}} = argmax\,(P_{x_{t}},\,axis =  0)$\;
    \For{c = 1 to C}{
    		$M_{P_{x_{t},c}} =  M_{P_{x_{t}}} [\,A_{P_{x_{t}}}== c\,] $\;
            $U_{c} = [\,U_{c},\, \,mean\,(M_{P_{x_{t},c}})\,] $\;
            $X_{t,c} = [\,X_{t,c},\, x_{t}\,]$\;
    }
 }
    \For{c = 1 to C}{
    	$X_{t,c,sort} = \,sort\,(X_{t,c} \,\,  w.r.t\,\,\,  U_{c}, \,descending\,\,order)$\;
         $ len_{th} = \,length\,(X_{t,c,sort}) \times (\textit{p}/C)\,\,\,\,\,   \rightarrow     (\textit{p}/C) \text{is the portion of class } \textit{c} $\;
         
         ${X^{'}}\textsubscript{t} = [\,{X^{'}}\textsubscript{t},\,X_{t,c,sort}\,[\,0 : len_{th}\, - \,1\, ] \,]$\;\;
          $\text{Calculate~\textit{h}\textsubscript{c}~for~each~class}$\;
          $x_{\textit{l}} = X_{t,c,sort} [\,len_{th} \, - \,1\,] $\;
          $P_{x_{\textit{l}}} = \textbf{\textit{f}}\,(w,x_{\textit{l}} )$\;
          $A_{P_{x_{\textit{l}}}} = argmax\,(P_{x_{\textit{l}}}, \,axis =  0)$\;
         $\textit{E}_{P_{x_{\textit{l}}}} = entropy\,(P_{x_{\textit{l}}})  \,\,\,\,\,   \rightarrow  \text{ normalized to }[0 ,1] $\;
         ${\textit{h}}_{c} = mean\,(\textit{ E}_{P_{x_{\textit{l}}}} [\,A_{P_{x_{\textit{l}}}} == c\,] )$ \;
         }
\Return ~${X^{'}}\textsubscript{t} , {\textit{h}}\textsubscript{c}$
\caption{Class-Based Sorting}
\label{image_selection_alg}
\end{algorithm}

\begin{figure}[t]
\begin{center}
\includegraphics[width=1\linewidth]{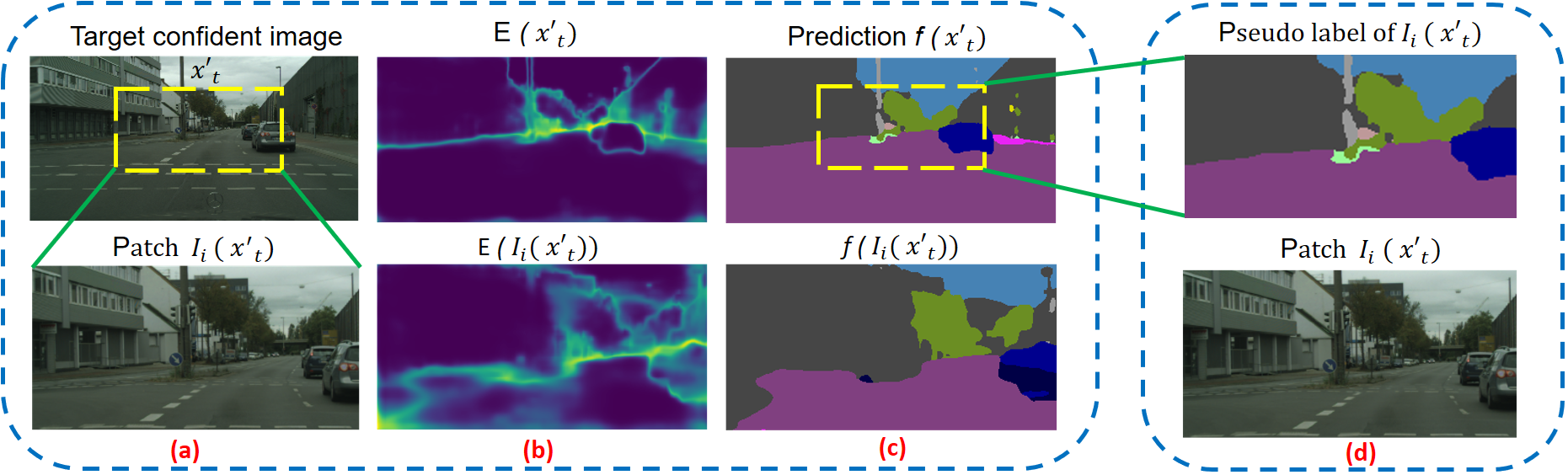}
\end{center}
   \caption{
   \textbf{Exploiting Scale-Invariance property for generated pseudo labels:}
    For an image $x^{t}$ belonging to target domain and its zoomed-in version scale-invariance property is violated.
    \textcolor{red}{(a)} Image $x^{t}$ and its extracted patch \textbf{\textit{I}}\textsubscript{i}. 
   \textcolor{red}{(b)} High self-entorpy values computed from the output probabilities indicate source model $\f$ is not confident about the labels assigned to resized patch.
   \textcolor{red}{(c)} comparison of the labels indicate violation of scale-invariance property 
   \textcolor{red}{(d)} Since original image exhibit low self-entropy we can use predictions over it as the pseudo-labels  for the resized patch.
    }

   
\label{fig:targ_sci}
\end{figure}

\subsection{Dynamic entropy threshold for class dependent filter selection}
The class based sorting takes in consideration all the pixels and does not make distinction between pixel-wise reliable and unreliable predictions. We define reliable or good predictions as by how low is the self-entropy of the prediction. If the entropy is low the prediction is more confident, if its high it means that the model is undecided which semantic label should be assigned to the pixel. 
Let, $P(x^{'}_{t})=\f(x^{'}_{t})$ be the predicted probability volume, and  $E _{x^{'}_{t}} =- \sum_c \,P_{c} (x'_{t})\,\,\textit{log}(P_{c} (x'_{t}))$ be entropy computed at each location. A binary filter map $F _{x^{'}_{t}}$ is generated by thresholding the entropy at every location, by a class specific threshold. 

\begin{equation}
F _{x^{'}_{t}}(h,w) = \left\{ \begin{array}{ll}
            1  & E_{x^{'}_{t}} (h,w)\leq h_{c}\text{~~;~~where~} c = argmax(P(x^{'}_{t})(h,w))\\
            0  &\text{otherwise}
        \end{array} \right.
\label{eq_cal_filter}
\end{equation}

Instead of being $h_c$ a global and constant hyper-parameter, $h_c$ is different for every class and depends upon predicted probabilities pixels belonging to that class in the selected confident set $X_{t}^{'}$. As the adaptation for that class improves the filter selection for that class becomes more tighter (Algo. \ref{image_selection_alg}).

\subsection{Self-Generated Scale-Invariant Examples}

Based on a reasonable assumption that a source domain consists of images with scene elements and objects of same class appearing in scale variations, we claim that model trained on such dataset should label same object with same semantic label regardless of its size in the image. We define this as \textit{scale-invariance property} of the model. 
As shown in Fig. \ref{fig:targ_sci} such a property is violated when target domain images are presented to the source model and could be used to guide the domain adaptation process. 
Specifically, lets assume $x^{'}_{t} \in {\textit{X$'$}}\textsubscript{t}$ be the one of the selected images, $F _{x^{'}_{t}}$ be the binary mask, and $P(x^{'}_{t}) = \f(x^{'}_{t})$ is the output probability volume.
Let $\textbf{\textit{R}}(x^{'}_{t},rec_i)$ be the operation applied on $x^{'}_{t}$ to extract $i^{th}$ patch from location $rec_i=(r_i, c_i, w_i, h_i)$ and resized to spatial size of $H \times W$. Then we can define,  $I_i^{t}=\textbf{\textit{R}}(x^{'}_{t},rec_i)$,
  $F_{x^{'}_{t}}^i=\textbf{\textit{R}}(F _{x^{'}_{t}},rec_i)$ and $P_{x^{'}_{t}}^i=\textbf{\textit{R}}(P _{x^{'}_{t}},rec_i)$ be the corresponding extracted and resized versions. 
We compute $\hat{y}_{t}^{i}$ is the one-hot encoded pseudo labels created from  $P_{x^{'}_{t}}^i$. Then loss for violating the scale invariance could be computed by Eq. \ref{eq_selflossExtracted}.
  
  \begin{equation}
\mathcal{L}_{seg}(I_i^{t}, \hat{y}_{t}^{i}) = - \sum_{h,w,c}^{H, W, C} F_t^{i,h,w} \hat{y}_t^{i,h,w,c} log(\f(I_i^{t})^{h,w,c}).
\label{eq_selflossExtracted}
\end{equation}

\begin{figure}[t!]
\begin{center}
\includegraphics[width=1\linewidth]{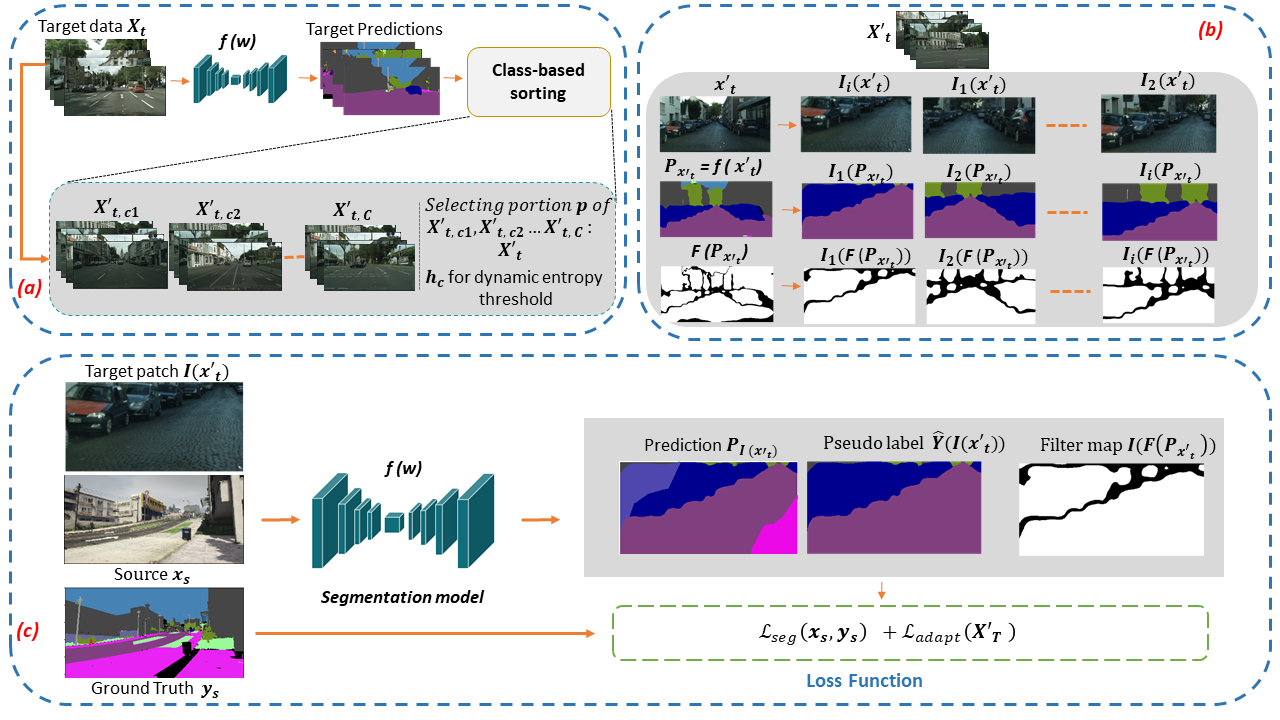}
\end{center}
   \caption{\textbf{Algorithm Overview:} Our algorithm consists of three main steps. \textcolor{red}{(a)} First, we have calculated the confidence of each target images \textit{X}\textsubscript{t} with reference to each class \textit{c}. We have sort out these images {X′}\textsubscript{t,c} of each class \textit{c} in descending order on the basis of their confidence value.  After that, we have selected the top portion from these sorted images \textit{X$'$}\textsubscript{t,c}  to form confident target data \textit{X$'$}\textsubscript{t}. \textcolor{red}{(b)} Second, we have extracted the random patches \textit{I}\textsubscript{i} from each confident images x$'$\textsubscript{t} of target domain \textit{X}\textsubscript{t}. These patches are the scale-invariant  with full-sized image. The model performs inconsistent on these patches and predict an output with high entropy prediction. To filter out the less confident pixels we have generated a filter map for each confident images \textit{x$'$}\textsubscript{t} by calculating their entropy with the help of threshold {h}\textsubscript{c} for each class {c}. \textcolor{red}{(c)} Third, we have trained the model by given loss function on these scale-invariant examples with their pseudo labels that are generated from the previous state of the model.
      }
\label{fig:intution_image}
\end{figure}

\subsection{Leveraging Focal Loss for Class-Imbalance}
Self-supervised approach for domain adaptation highly dependent on information represented in selected confident images of the target domain.
Biased distribution, i.e. number of pixels per class, in the road scenes creates a class imbalance problem.
Even after the class based sorting (Sec.~\ref{sec:classSort}) and class dependent entropy thresholding, classes with high volume of pixels in target dataset (such as  road, building, vegetation, etc.) end up having more contribution towards loss function. 
Classes which appear infrequently and/or have less volume of pixels per image will contribute less and hence adaptation will be slow. 
To eliminate the effect of class imbalance problem, we incorporate the focal loss \cite{lin2017focal}, so that cross-entropy function of each pixel is weighted by the 
based on pixel confidence.
Focal loss balanced the loss for each pixel based on their confident level. This approach of applying focal loss balance the learning process of self-supervised learning equally to each class.  In this work, we apply focal loss during the training of scale-invariant examples. Eq. \ref{eq_FL} shows the formulation of focal loss. 


\begin{equation}
\mathcal{L}_{FL}( I_i^{t}, \hat{y}_{t}^{i} ) = - \sum_{h,w,c}^{H, W, C} \,\hat{y}_t^{i,h,w,c}\,\,\textbf{\textit{log}}(\f(I_i^{t})^{h,w,c}) ( 1 - \f(I_i^{t})^{h,w,c})^\gamma \\
\label{eq_FL}
\end{equation}


Where $\gamma$ is the hyperparameter that controls the focus and generally have value between 0 to 5. Low value bring it closer cross-entropy and high value focusing only on the hard examples. We set $\gamma$ to middle value,3. 
\subsection{Adaptation}

During adaptation, for each round $r$, we perform class based sorting of target dataset to create subset $X^{'}_{T}$. For each $x^{'}_{t} \in X^{'}_{T}$,  $k$ patches are extracted. 
Out total loss is defined as 
\begin{equation}
 \mathcal{L}_{LSE} = \sum_{x_s \in X_S} \mathcal{L}_{seg}( x_{s},y_{s}) +\mathcal{L}_{adapt}(X^{'}_{T})
\label{eq_final}
\end{equation}
where first term is cross entropy loss over source domain {X}\textsubscript{s} to prevent the model from forgetting the previous knowledge. Second term, is adaptation loss computed as summation of focal loss Eq. \ref{eq_FL} and segmentation loss  (Eq. \ref{eq_selflossExtracted}), trying to minimize loss of violating scale-invariance.

\begin{equation}
 \mathcal{L}_{adapt}(X^{'}_{T}) = \sum_{x^{'}_{t} \in X^{'}_{T}} \sum_{i}^{k} 
 \beta \mathcal{L}_{FL}( I_i^{t}, \hat{y}_{t}^{i} )
 + \mathcal{L}_{seg}(I_i^{t}, \hat{y}_{t}^{i}),
\end{equation}
$\beta$ is a hyperparameter that controls the effect of focal loss on self-supervised domain adaptation.
In the end, we adapt the model with an iterative process for each rounds \textit{r}. 
Fig. \ref{fig:intution_image} shows complete model.
\section{Experiments and Results} 
In this section, we provide implementation details and experimental setup of our proposed approach. We evaluate the proposed self-supervised learning strategy on standard synthetic to real domain adaptation setup and present a detailed comparison with state-of-the-art methods.

\subsection{Experimental Details}
\textbf{Network Architecture:}
For a fair comparison we follow the standard practice of using FCN-8s \cite{long2015fully} with VGG16 and DeepLab-v2 \cite{chen2017deeplab} with ResNet-101 \cite{he2016deep} as our baseline approaches. 
We have used pretrained models for further adaptation towards the target domain

\textbf{Datasets and Evaluation Metric:} 
To evaluate the proposed approach, we have used benchmark synthetic datasets, e.g., GTA5 \cite{richter2016playing} and SYNTHIA-RAND-CITYSCAPES  \cite{ros2016synthia} as our source domain datasets and real imagery Cityscapes\cite{cordts2016cityscapes} as our target domain dataset.
The GTA5 dataset consists of 24966 high resolution (1052 x 1914) densely annotated images captured from the GTA5 game. 
Similarly, SYNTHIA contains 9400 labeled images with a spatial resolution of 760 x 1280. 
The Cityscapes datasets has 2975 training images and 500 validation images. 
We use mean intersection over union (mIoU) as the evaluation metric and evaluate the proposed approach on compatible 19 and 16 classes for GTA to Cityscapes and SYNTHIA to Cityscapes adaptation respectively. Due to GPU memory limitations we use the highest spatial size of $512 \times 1024$.
 
\textbf{Implementation Details:}
We have used PyToch deep learning framework to implement our algorithm with a Tesla k80 GPU having 12GB of memory.
To select number of high confident images for each class, we choose $p = 0.1$ and after each round increment it with 0.05. $k=4$ number of patches, of spatial size of $256 \times 512$, are chosen randomly and resized to $512 \times 1024$. For focal loss, we use $\gamma = 3$ and $\beta = 0.1$ in-order to focus on hard examples. We used Adam optimizer \cite{kingma2014adam} with learning  rate and momentum of 1x10\textsuperscript{-6} and 0.9 respectively.

\subsection{Comparisons with state-of-the-art Methods}
To compare with existing methods, we perform  experiments of adapting to Cityscapes from two different synthetic datasets, GTA5 and SYNTHIA. All experiments were done under the standard settings.

\textbf{GTA5 to Cityscapes:} Table \ref{gta5tocityscape} shows the comparison of our result with existing state of the art domain adaptation methods in semantic segmentation from GTA5 
to Cityscapes
respectively. 
Proposed approach reports state-of-the-art results on VGG16-FCN8 \cite{long2015fully} and ResNet101 \cite{he2016deep}, for self-training based adaptation methods. It outperforms most of the non self-training methods and complex methods too, and is comparative to state-of-the-art. 
 We report the results with and without the focal loss to see the effect on the model regarding class balance adaptation.
Due to focal loss, the small/infrequent objects benefit specifically.



\begin{table}[t]
  \centering
  \def\arraystretch{1.5}%
  \caption{Results from GTA5 to Cityscapes. We report the results of our algorithm by presenting IoU of each class and also overall mIoU. `V' and \textcolor{blue}{`R'} represents VGG-FCN8 and \textcolor{blue}{ResNet101} as our baseline network. `ST' and `AT' represents self-training and adversarial training respectively. We report the best results in \textbf{bold}.}
  \resizebox{12cm}{!}{
    \begin{tabular}{c|c|c|ccccccccccccccccccc|c}
    
    \specialrule{2pt}{0.5pt}{0.5pt}
    \multicolumn{22}{c}{\textbf{GTA5 to Cityscapes}} \\
    \specialrule{2pt}{0.5pt}{0.5pt}
    
          & Arch.    &  Meth.  & 
            road  &   sidewalk  &   building  &   wall  &   fence  &   pole  &   light  &   sign  &   veg  &   terrain  &   sky  &   person  &   rider  &   car  &   truck  &   bus  &   train  &   mbike  &   bike  & mIoU \\
    \specialrule{2pt}{0.5pt}{0.5pt}
    
    FCN wild \cite{hoffman2016fcns} & V & AT    & 70.4  & 32.4  & 62.1  & 14.9  & 5.4   & 10.9  & 14.2  & 2.7   & 79.2  & 21.3  & 64.6  & 44.1  & 4.2   & 70.4  & 8.0     & 7.3   & 0.0     & 3.5   & 0.0     & 27.1 \\\hline
    CyCADA \cite{hoffman2017cycada}& V & AT    & 85.2  & \textbf{37.2}  & 76.5  & 21.8  & 15.0    & 23.8  & 22.9  & \textbf{21.5} & 80.5  & 31.3  & 60.7  & 50.5  & 9.0     & 76.9  & 17.1  & 28.2  & 4.5   & 9.8   & 0.0     & 35.4 \\\hline
    
    ROAD \cite{chen2018road}& V  & AT    & 85.4  & 31.2  & 78.6  & 27.9  & 22.2& 21.9  & 23.7  & 11.4  & 80.7  & 29.3  & 68.9  & 48.5  & 14.1  & 78.0    & 19.1  & 23.8  & 9.4   & 8.3   & 0.0     & 35.9 \\
    
 & \textcolor{blue}{R} & \textcolor{blue}{AT} & \textcolor{blue}{76.3} &  \textcolor{blue}{36.1} &  \textcolor{blue}{69.6} &  \textcolor{blue}{28.6} &  \textcolor{blue}{22.4} &  \textcolor{blue}{28.6} &  \textcolor{blue}{29.3} &  \textcolor{blue}{14.8}  & \textcolor{blue}{82.3} &  \textcolor{blue}{35.3}  & \textcolor{blue}{72.9}  & \textcolor{blue}{54.4}  & \textcolor{blue}{17.8}  & \textcolor{blue}{78.9}  & \textcolor{blue}{27.7} &  \textcolor{blue}{30.3}  &   \textcolor{blue}{4.0} &  \textcolor{blue}{24.9 } & \textcolor{blue}{12.6} &  \textcolor{blue}{39.4} \\ \hline
 CLAN \cite{luo2019taking}& V & AT    & 88.0    & 30.6  & 79.2  & 23.4  & 20.5  & 26.1  & 23.0    & 14.8  & 81.6  & \textbf{34.5 } & 72.0    & 45.8  & 7.9   & 80.5  & \textbf{26.6} & 29.9 & 0.0     & 10.7  & 0.0     & 36.6\\
    & \textcolor{blue}{R} & \textcolor{blue}{AT}   &   \textcolor{blue}{87.0}  & \textcolor{blue}{27.1} &  \textcolor{blue}{79.6} &  \textcolor{blue}{27.3} &  \textcolor{blue}{23.3} &  \textcolor{blue}{28.3}  & \textcolor{blue}{35.5}  & \textcolor{blue}{24.2} &  \textcolor{blue}{83.6} &  \textcolor{blue}{27.4} &  \textcolor{blue}{74.2} &  \textcolor{blue}{58.6} &  \textcolor{blue}{28.0}  & \textcolor{blue}{76.2 } & \textcolor{blue}{33.1 }&  \textcolor{blue}{36.7 }&  \textcolor{blue}{6.7 }&  \textcolor{blue}{\textbf{31.9}} &  \textcolor{blue}{31.4} &  \textcolor{blue}{43.2} \\\hline 
    Curr. DA \cite{zhang2017curriculum}& V & AT    & 74.9 &  22.0 &  71.7 &  6.0 &  11.9 &  8.4 &  16.3 &  11.1 &  75.7 &  13.3  & 66.5 &  38.0 &  9.3  & 55.2  & 18.8 &  18.9 &  0.0 &  16.8 &  14.6 &  28.9  \\ 
    \specialrule{1.2pt}{0.5pt}{0.5pt}
    AdvEnt \cite{vu2018advent}& V  & AT,ST    & 86.9  & 28.7  & 78.7  & 28.5  & 25.2& 17.1  & 20.3  & 10.9  & 80.0  & 26.4  & 70.2  & 47.1  & 8.4  & 81.5    & 26.0  & 17.2  & \textbf{18.9}   & 11.7   & 1.6     & 36.1 \\
    & \textcolor{blue}{R} & \textcolor{blue}{AT,ST} & \textcolor{blue}{89.4} &  \textcolor{blue}{33.1} &  \textcolor{blue}{81.0} &  \textcolor{blue}{26.6} &  \textcolor{blue}{26.8} &  \textcolor{blue}{27.2} &  \textcolor{blue}{33.5} &  \textcolor{blue}{24.7}  & \textcolor{blue}{83.9} &  \textcolor{blue}{36.7}  & \textcolor{blue}{78.8}  & \textcolor{blue}{58.7}  & \textcolor{blue}{30.5}  & \textcolor{blue}{84.8}  & \textcolor{blue}{\textbf{38.5}} &  \textcolor{blue}{44.5}  & \textcolor{blue}{1.7} &  \textcolor{blue}{31.6 } & \textcolor{blue}{32.4} &  \textcolor{blue}{45.5}
    \\\hline
    SSF-DAN \cite{du2019ssf}& V & ST,AT & \textbf{88.7}  & 32.1  & 79.5  & 29.9  & 22.0    & 23.8  & 21.7  & 10.7  & 80.8  & 29.8  & 72.5  & 49.5  & 16.1  & \textbf{82.1}  & 23.2  & 18.1  & 3.5   & \textbf{24.4}  & 8.1   & 37.7 
    \\
    & \textcolor{blue}{R} & \textcolor{blue}{ST,AT}    &  \textcolor{blue}{90.3}  & \textcolor{blue}{38.9} &  \textcolor{blue}{81.7 }&  \textcolor{blue}{24.8 } & \textcolor{blue}{22.9 } & \textcolor{blue}{30.5 }&  \textcolor{blue}{37.0 } & \textcolor{blue}{21.2 }&  \textcolor{blue}{84.8 }&  \textcolor{blue}{\textbf{38.8} }&  \textcolor{blue}{76.9 } & \textcolor{blue}{58.8 } & \textcolor{blue}{30.7 }&  \textcolor{blue}{\textbf{85.7} }&  \textcolor{blue}{30.6 } & \textcolor{blue}{38.1 }&  \textcolor{blue}{5.9  }& \textcolor{blue}{28.3 }&  \textcolor{blue}{36.9 } & \textcolor{blue}{45.4}
\\\specialrule{1.2pt}{0.5pt}{0.5pt}

    CBST \cite{zou2018unsupervised}& V  & ST    & 66.7  & 26.8  & 73.7  & 14.8  & 9.5   & 28.3& 25.9  & 10.1  & 75.5  & 15.7  & 51.6  & 47.2  & 6.2   & 71.9  & 3.7   & 2.2   & 5.4   & 18.9  & \textbf{32.4 }& 30.9 \\\hline

    PyCDA\cite{lian2019constructing}& V   & ST & 86.7 & 24.8 &\textbf{ 80.9} & 21.4 &\textbf{ 27.3}  & \textbf{30.2}  & 26.6    & 21.1  & \textbf{86.6} & 28.9 & 58.8 & \textbf{53.2 }& 17.9 & 80.4& 18.8  & 22.4    & 4.1   & 9.7 & 6.2  & 37.2 \\
    & 
    \textcolor{blue}{R }& \textcolor{blue}{ST }   &  \textcolor{blue}{\textbf{90.5}}  & \textcolor{blue}{36.3} &  \textcolor{blue}{\textbf{84.4} }&  \textcolor{blue}{\textbf{32.4} } & \textcolor{blue}{\textbf{28.7} }&  \textcolor{blue}{\textbf{34.6} } & \textcolor{blue}{36.4 } & \textcolor{blue}{31.5 }&  \textcolor{blue}{\textbf{86.8} } & \textcolor{blue}{37.9 } & \textcolor{blue}{78.5 } & \textcolor{blue}{\textbf{62.3} } & \textcolor{blue}{21.5 }&  \textcolor{blue}{85.6 }&  \textcolor{blue}{27.9 } & \textcolor{blue}{34.8 }&  \textcolor{blue}{\textbf{18.0} } & \textcolor{blue}{22.9} &  \textcolor{blue}{\textbf{49.3} }&  \textcolor{blue}{47.4}\\
    \specialrule{1.4pt}{0.5pt}{0.5pt}
    LSE   & V &  ST   &  80.2  & 26.6 & 78.1  & 28.4  & 17.3  & 19.8  & 27.6 & 12.2  & 78.6  & 23.6 & 72.0  & 50.8  & 14.8  & 81.2 & 22.5  & 20.3  & 4.0 & 20.1  & 14.5  & 36.4 \\
    
     LSE + FL & V    & ST   &  86.0 & 26.0  & 76.7  & \textbf{33.1}  & 13.2  & 21.8  & \textbf{30.1} & 16.5  & 78.8  & 25.8  & \textbf{74.7}  & 50.6  &\textbf{ 18.7}  & 81.8  & 22.5  & \textbf{30.5}  & 12.3 & 16.9  & 25.4  &\textbf{39.0} \\
     
     \textcolor{blue}{LSE + FL}&
      \textcolor{blue}{R}&
      \textcolor{blue}{ST}&
      \textcolor{blue}{90.2}  & \textcolor{blue}{\textbf{40.0}} &  \textcolor{blue}{83.5}&  \textcolor{blue}{31.9 } & \textcolor{blue}{26.4 }&  \textcolor{blue}{32.6 } & \textcolor{blue}{\textbf{38.7} } & \textcolor{blue}{\textbf{37.5} }&  \textcolor{blue}{81.0 } & \textcolor{blue}{34.2} & \textcolor{blue}{\textbf{84.6} } & \textcolor{blue}{61.6 } & \textcolor{blue}{\textbf{33.4}}&  \textcolor{blue}{82.5}&  \textcolor{blue}{32.8} & \textcolor{blue}{\textbf{45.9} }&  \textcolor{blue}{6.7 } & \textcolor{blue}{29.1} &  \textcolor{blue}{30.6 }&  \textcolor{blue}{\textbf{47.5}}\\
   \hline
    \end{tabular}

    }
  \label{gta5tocityscape}%
\end{table}

\textbf{SYNTHIA to Cityscapes:} Table \ref{syntoCity} describes the quantitative results of LSE and a detailed comparison with existing methods.
Like previous methods \cite{Iqbal_2020_WACV}, we report both the mIoU (16 classes) and mIoU* (13 classes) for the classes compatible with Cityscapes. The LSE+FL performs comparative to other complex methods based on adversarial learning, however, in self-training setting LSE+FL shows 4.1\% mIoU gain over state-of-the-art PyCDA \cite{zhao2017pyramid}.\\

  \begin{table*}[t]
  
  \centering
  \def\arraystretch{1.2}%
  \caption{mIoU (16-categories) and mIoU* (13-categories) results from SYNTHIA to Cityscapes. `V' and \textcolor{blue}{`R'} represent VGG-FCN8 and \textcolor{blue}{ResNet101} as our baseline network. `ST' and `AT' represent self-training and adversarial training, respectively. We have reported the highest results in \textbf{bold}.}
  \resizebox{12cm}{!}{
    \begin{tabular}{c|c|c|cccccccccccccccc|c|c}
    \specialrule{2pt}{0.5pt}{0.5pt}
    \multicolumn{20}{c}{SYNTHIA to Cityscapes} \\
    \specialrule{2pt}{0.5pt}{0.5pt}
          & %
          Arch.
          &   Meth.  &  road  &  sidewalk  &  building  &  wall  &  fence  &  pole  &  light  &  sign  &  veg  &  sky  &  person  &  rider  &  car  &  bus  &  mbike  &  bike  & mIoU  & mIoU* \\
          \specialrule{2pt}{0.5pt}{0.5pt}
    ROAD \cite{chen2018road}  &V& AT    & 77.7  & 30.0    & 77.5  & 9.6   & 0.3   & \textbf{25.8}  & 10.3  & 15.6  & 77.6  & 79.8  & 44.5  & 16.6  & 67.8  & 14.5  & 7.0     & 23.8  & 36.2  & -  \\\hline
    
    CLAN \cite{luo2019taking} &V& AT    & 80.4  & 30.7  & 74.7  & -     & -     & -     & 1.4   & 8.0     & 77.1  & 79.0    & 46.5  & 8.9   & 73.8  & 18.2  & 2.2   & 9.9   & -     & 39.3 \\
    & \textcolor{blue}{R} & \textcolor{blue}{AT} &  \textcolor{blue}{81.3} &  \textcolor{blue}{37.0}  & \textcolor{blue}{80.1} &  \textcolor{blue}{- }&  \textcolor{blue}{-} &  \textcolor{blue}{-} &  \textcolor{blue}{16.1}  & \textcolor{blue}{13.7}  & \textcolor{blue}{78.2} &  \textcolor{blue}{81.5} &  \textcolor{blue}{53.4} &  \textcolor{blue}{21.2} &  \textcolor{blue}{73.0} &  \textcolor{blue}{32.9}  & \textcolor{blue}{22.6} &  \textcolor{blue}{30.7}  & \textcolor{blue}{-}  & \textcolor{blue}{47.8} \\\hline 
    
    Curr. DA \cite{zhang2017curriculum} &V& AT    & 65.2  & 26.1  & 74.9  & 0.1   & 0.5   & 10.7  & 3.7   & 3.0     & 76.1  & 70.6  & 47.1  & 8.2   & 43.2  & 20.7  & 0.7   & 13.1  & -    & 34.8 \\\specialrule{1.2pt}{0.5pt}{0.5pt}
    AdvEnt \cite{vu2018advent} &V& AT,ST    & 67.9  & 29.4  & 71.9  & 6.3     & 0.3     & 19.9     & 0.6   & 2.6     & 74.9  & 74.9     & 35.4  & 9.6   & 67.8  & 21.4  & 4.1   & 15.5   & 31.4     & 36.6 \\
    & \textcolor{blue}{R} & \textcolor{blue}{AT,ST} &  \textcolor{blue}{85.6} &  \textcolor{blue}{42.2}  & \textcolor{blue}{79.7} &  \textcolor{blue}{8.7 }&  \textcolor{blue}{0.4} &  \textcolor{blue}{25.9} &  \textcolor{blue}{5.4}  & \textcolor{blue}{8.1}  & \textcolor{blue}{80.4} &  \textcolor{blue}{84.1} &  \textcolor{blue}{57.9} &  \textcolor{blue}{23.8} &  \textcolor{blue}{73.3} &  \textcolor{blue}{36.4}  & \textcolor{blue}{14.2} &  \textcolor{blue}{33.0}  & \textcolor{blue}{41.2}  & \textcolor{blue}{48.0} \\\hline
    SSF-DAN \cite{du2019ssf} &V& ST,AT & \textbf{87.1}  & 36.5  & \textbf{79.7 } & -     & -     & -     & 13.5  & 7.8   & 81.2  & 76.7  & 50.1  & 12.7  & \textbf{78.0 }   & \textbf{35.0}   & 4.6   & 1.6   & -     & 43.4 \\
    & \textcolor{blue}{R} & \textcolor{blue}{ST,AT} &  \textcolor{blue}{ \textbf{84.6}}  & \textcolor{blue}{41.7 }&  \textcolor{blue}{80.8 }& \textcolor{blue}{-}&\textcolor{blue}{-}&\textcolor{blue}{-}& \textcolor{blue}{11.5} &  \textcolor{blue}{14.7} &  \textcolor{blue}{80.8 }&  \textcolor{blue}{\textbf{85.3} }&  \textcolor{blue}{57.5 }&  \textcolor{blue}{21.6 }&  \textcolor{blue}{82.0 }&  \textcolor{blue}{36.0 }&  \textcolor{blue}{19.3  }& \textcolor{blue}{\textbf{34.5} } & \textcolor{blue}{-} & \textcolor{blue}{50.0}
\\\specialrule{1.2pt}{0.5pt}{0.5pt}
    CBST \cite{zou2018unsupervised} &V& ST    & 69.6  & 28.7  & 69.5  &\textbf{ 12.1}  & 0.1   & 25.4  & 11.9  & 13.6  & \textbf{82.0}    & \textbf{81.9 } & 49.1  & 14.5  & 66    & 6.6   & 3.7   & \textbf{32.4 } & 35.4  & 36.1 \\\hline

    PyCDA\cite{lian2019constructing}  &V& ST & 80.6    & 26.6  & 74.5  & 2.0   & 0.1   & 18.1  & 13.7  & 14.2  & 80.8  & 71.0    & 48.0  & \textbf{19.0}  & 72.3  & 22.5    & \textbf{12.1}   & 18.1  & 35.9    & 42.6  \\
    & \textcolor{blue}{R} & \textcolor{blue}{ST} &   \textcolor{blue}{75.5} &  \textcolor{blue}{30.9} &  \textcolor{blue}{\textbf{83.3}} &  \textcolor{blue}{\textbf{20.8}} &  \textcolor{blue}{\textbf{0.7} }&  \textcolor{blue}{\textbf{32.7}}  & \textcolor{blue}{\textbf{27.3}}  & \textcolor{blue}{\textbf{33.5}}  & \textcolor{blue}{\textbf{84.7}} &  \textcolor{blue}{85.0}  & \textcolor{blue}{\textbf{64.1}} &  \textcolor{blue}{25.4}  & \textcolor{blue}{\textbf{85.0}}  & \textcolor{blue}{\textbf{45.2}}  & \textcolor{blue}{21.2} &  \textcolor{blue}{32.0}  & \textcolor{blue}{\textbf{46.7}} &  \textcolor{blue}{\textbf{53.3}}\\
    \specialrule{1.2pt}{0.5pt}{0.5pt}
    LSE  &V &  ST  &  82.2     &  38.4     &   79.0    &   2.2    &   0.5    &     25.3  &9.6 &  20.7& 78.6     &   77.4    &    51.7   &    18.0   &  72.9     &  21.7     &  11.1     &   22.2    &  38.2        &44.9  \\
LSE + FL  &V &  ST   &    83.6   &     \textbf{39.6 }&   79.3    &  3.6     &  \textbf{ 0.9}    &   25.3    &  \textbf{14.1 }   & \textbf{ 26.1}     &    79.4   &     76.5  &   \textbf{51.0 }   &   18.1    &  75.7     &  22.5     &    12.0   &   32.1    &  \textbf{40.0}     & \textbf{47.0} \\
    \textcolor{blue}{LSE + FL} & \textcolor{blue}{R} & \textcolor{blue}{ST} &   \textcolor{blue}{82.9} &  \textcolor{blue}{\textbf{43.1}} &  \textcolor{blue}{78.1} &  \textcolor{blue}{9.3} &  \textcolor{blue}{0.6 }&  \textcolor{blue}{28.2}  & \textcolor{blue}{9.1}  & \textcolor{blue}{14.4}  & \textcolor{blue}{77.0} &  \textcolor{blue}{83.5}  & \textcolor{blue}{58.1} &  \textcolor{blue}{\textbf{25.9}}  & \textcolor{blue}{71.9}  & \textcolor{blue}{38.0}  & \textcolor{blue}{\textbf{29.4}} &  \textcolor{blue}{31.2}  & \textcolor{blue}{42.6} &  \textcolor{blue}{49.4}\\

    \hline
    \end{tabular}
    }
  \label{syntoCity}
  \end{table*}

\subsection{Analysis }
\begin{figure}
\begin{center}
\includegraphics[width=1\linewidth]{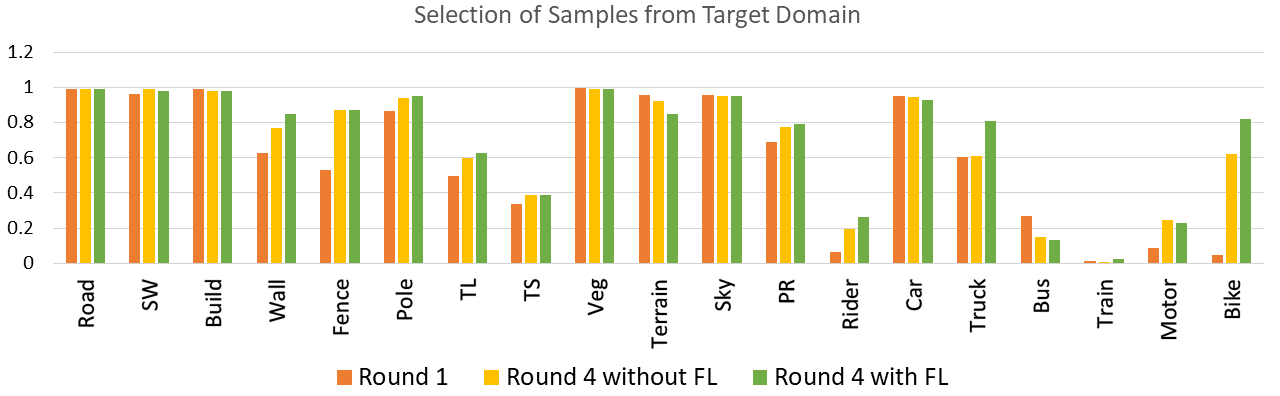}
\end{center}
   \caption{Effect of focal loss on each class after the first and the fourth round of domain adaptation with self-supervised learning for semantic segmentation, evaluated for GTA5 to Cityscape with VGG16-FCN8 baseline network.}
\label{effect_of_focal_loss}
\end{figure}

To demonstrate the reasoning of the working principle for the proposed algorithm, we evaluate different aspect of our algorithm. \\\\
\textbf{Effect of Focal Loss:}
To verify the effect of focal loss on each class equally, we calculate the number of images selected for each class after a few rounds. Focal loss can affect the smaller classes for each class on different rounds, as shown in Figure \ref{effect_of_focal_loss}. The graph demonstrates the effect on different classes to balance the effect of learning for self-supervised domain adaptation. For each class, the 
Figure \ref{effect_of_focal_loss} shows three bars, red shows the number of images selected on the first round of adaptation, whereas the orange and green are the corresponding values of selected images after fourth round and with and without focal loss respectively. It can be seen that the focal loss balances the selection process especially for infrequent classes, by maximizing their prediction probabilities.

\textbf{Performance Gap:} We also compare the performance of our algorithm using the performance gap with other state-of-the-art methods of domain adaptation. Table \ref{perf_gap} shows the performance gap of different algorithms with their oracle values. Our algorithm clearly shows the best results with a gap \textbf{-21.3} as compared to other algorithms we mentioned.\\\\

\capbtabbox{%

 \begin{tabular}{c|c|c|c}
    \specialrule{1.2pt}{0.5pt}{0.5pt}
    \multicolumn{4}{c}{Performance Table } \\
    \specialrule{1.2pt}{0.5pt}{0.5pt}
    \multicolumn{4}{c}{GTA5 to Cityscapes (VGG16-FCN8)} \\
    \specialrule{1.2pt}{0.5pt}{0.5pt}
    Method & Oracle  & mIoU \% & gap (\%) \\
    \specialrule{1.2pt}{0.5pt}{0.5pt}
    FCN wild \cite{hoffman2016fcns} & 64.6  & 27.1  & -37.5 \\

    CyCADA \cite{hoffman2017cycada} & 60.3  & 35.4  & -24.9 \\
    ROAD\cite{chen2018road}  & 64.6  & 35.9  & -28.7 \\
    CLAN\cite{luo2019taking}  & 64.6  & 36.6  & -28.0 \\
    AdvEnt \cite{vu2018advent} & 61.8  & 36.1  & -25.7 \\
    SSF-DAN\cite{du2019ssf} & 65.1  & 37.7  & -27.4 \\
    CBST \cite{zou2018unsupervised} & 65.1  & 30.9  & -34.2 \\
    PyCDA\cite{lian2019constructing} & 65.1  & 37.2  & -27.9 \\
    \specialrule{1.2pt}{0.5pt}{0.5pt}
    Ours  & 60.3  & 39.9  & \textbf{-21.3} \\
        \end{tabular}%
}{%
  \caption{Comparisons of performance gap of adaptation algorithms vs oracle scores}%
  \label{perf_gap}
}

\begin{figure}[t]
\begin{center}
\includegraphics[width=1.0\linewidth]{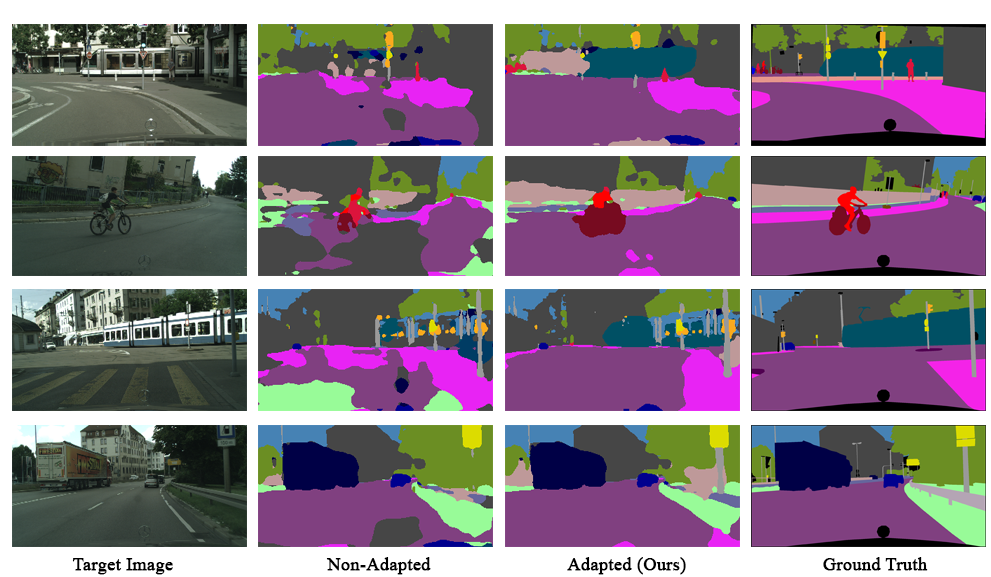}
\end{center}
   \caption{Qualitative results of our algorithm with self-supervised domain adaptation for GTA5 to Cityscapes. For each example, we show images without adaptation and with adaptation as our result. We also show the ground truth for each image.   
}
\label{algorithm_flow_fig}
\end{figure}
 


    %
 %

\section{Conclusion}
In this paper, we have proposed a novel approach of self-supervised domain adaptation method by exploiting the scale-invariance properties of the semantic segmentation model. 
In general images in dataset, especially road-scene dataset, contains objects in varying sizes and scene elements closer and far away from the. 
The scale invarance property of the model is defined as ability to assign same semantic labels to scaled instance of the image or parts of image as it will assign to the original image. In simple words regardless of size variation of object it should be similarly semantically labeled. 
We show that for the target domain this property is violated and could be used to direct the adaptation label by using the pseudo-labels for the original size images as pseudo-labels for the zoomed in region. 
Multiple strategies were employed to counter the class imbalance problem and  pseudo-label selection problem. 
Class specific sorting algorithm is desinged to select images from target dataset such that all classes are equally represented at image level. Dynamic class dependent entropy threshold mechanism is presented to allow classes at different levels of adaptation have different threshold. 
Finally, a focal loss is introduced to guide the adaptation process.
Our experimenal results are competitive to state-of-the-ar algorithms and outpeform state-of-the-art self-training methods. 
\clearpage
%

%
\bibliographystyle{splncs04}
\bibliography{egbib}

\begin{thebibliography}{10}
\providecommand{\url}[1]{\texttt{#1}}
\providecommand{\urlprefix}{URL }
\providecommand{\doi}[1]{https://doi.org/#1}

\bibitem{badrinarayanan2017segnet}
Badrinarayanan, V., Kendall, A., Cipolla, R.: Segnet: A deep convolutional
  encoder-decoder architecture for image segmentation. IEEE transactions on
  pattern analysis and machine intelligence  \textbf{39}(12),  2481--2495
  (2017)

\bibitem{chen2017deeplab}
Chen, L.C., Papandreou, G., Kokkinos, I., Murphy, K., Yuille, A.L.: Deeplab:
  Semantic image segmentation with deep convolutional nets, atrous convolution,
  and fully connected crfs. IEEE transactions on pattern analysis and machine
  intelligence  \textbf{40}(4),  834--848 (2017)

\bibitem{chen2018deeplab}
Chen, L.C., Papandreou, G., Kokkinos, I., Murphy, K., Yuille, A.L.: Deeplab:
  Semantic image segmentation with deep convolutional nets, atrous convolution,
  and fully connected crfs. IEEE transactions on pattern analysis and machine
  intelligence  \textbf{40}(4),  834--848 (2018)

\bibitem{chen2017no}
Chen, Y.H., Chen, W.Y., Chen, Y.T., Tsai, B.C., Frank~Wang, Y.C., Sun, M.: No
  more discrimination: Cross city adaptation of road scene segmenters. In:
  Proceedings of the IEEE International Conference on Computer Vision. pp.
  1992--2001 (2017)

\bibitem{chen2018road}
Chen, Y., Li, W., Van~Gool, L.: Road: Reality oriented adaptation for semantic
  segmentation of urban scenes. In: Proceedings of the IEEE Conference on
  Computer Vision and Pattern Recognition. pp. 7892--7901 (2018)

\bibitem{cordts2016cityscapes}
Cordts, M., Omran, M., Ramos, S., Rehfeld, T., Enzweiler, M., Benenson, R.,
  Franke, U., Roth, S., Schiele, B.: The cityscapes dataset for semantic urban
  scene understanding. In: Proceedings of the IEEE conference on computer
  vision and pattern recognition. pp. 3213--3223 (2016)

\bibitem{du2019ssf}
Du, L., Tan, J., Yang, H., Feng, J., Xue, X., Zheng, Q., Ye, X., Zhang, X.:
  Ssf-dan: Separated semantic feature based domain adaptation network for
  semantic segmentation. In: Proceedings of the IEEE International Conference
  on Computer Vision. pp. 982--991 (2019)

\bibitem{ganin2016domain}
Ganin, Y., Ustinova, E., Ajakan, H., Germain, P., Larochelle, H., Laviolette,
  F., Marchand, M., Lempitsky, V.: Domain-adversarial training of neural
  networks. The Journal of Machine Learning Research  \textbf{17}(1),
  2096--2030 (2016)

\bibitem{he2016deep}
He, K., Zhang, X., Ren, S., Sun, J.: Deep residual learning for image
  recognition. In: Proceedings of the IEEE conference on computer vision and
  pattern recognition. pp. 770--778 (2016)

\bibitem{hoffman2017cycada}
Hoffman, J., Tzeng, E., Park, T., Zhu, J.Y., Isola, P., Saenko, K., Efros,
  A.A., Darrell, T.: Cycada: Cycle-consistent adversarial domain adaptation.
  arXiv preprint arXiv:1711.03213  (2017)

\bibitem{hoffman2016fcns}
Hoffman, J., Wang, D., Yu, F., Darrell, T.: Fcns in the wild: Pixel-level
  adversarial and constraint-based adaptation. arXiv preprint arXiv:1612.02649
  (2016)

\bibitem{Iqbal_2020_WACV}
Iqbal, J., Ali, M.: Mlsl: Multi-level self-supervised learning for domain
  adaptation with spatially independent and semantically consistent labeling.
  In: Proceedings of the IEEE/CVF Winter Conference on Applications of Computer
  Vision (WACV) (March 2020)

\bibitem{kingma2014adam}
Kingma, D.P., Ba, J.: Adam: A method for stochastic optimization. arXiv
  preprint arXiv:1412.6980  (2014)

\bibitem{laine2016temporal}
Laine, S., Aila, T.: Temporal ensembling for semi-supervised learning. arXiv
  preprint arXiv:1610.02242  (2016)

\bibitem{lian2019constructing}
Lian, Q., Lv, F., Duan, L., Gong, B.: Constructing self-motivated pyramid
  curriculums for cross-domain semantic segmentation: A non-adversarial
  approach. In: Proceedings of the IEEE International Conference on Computer
  Vision. pp. 6758--6767 (2019)

\bibitem{lin2017focal}
Lin, T.Y., Goyal, P., Girshick, R., He, K., Doll{\'a}r, P.: Focal loss for
  dense object detection. In: Proceedings of the IEEE international conference
  on computer vision. pp. 2980--2988 (2017)

\bibitem{long2015fully}
Long, J., Shelhamer, E., Darrell, T.: Fully convolutional networks for semantic
  segmentation. In: Proceedings of the IEEE conference on computer vision and
  pattern recognition. pp. 3431--3440 (2015)

\bibitem{luo2019taking}
Luo, Y., Zheng, L., Guan, T., Yu, J., Yang, Y.: Taking a closer look at domain
  shift: Category-level adversaries for semantics consistent domain adaptation.
  In: Proceedings of the IEEE Conference on Computer Vision and Pattern
  Recognition. pp. 2507--2516 (2019)

\bibitem{quinonero2008covariate}
Qui{\~n}onero-Candela, J., Sugiyama, M., Schwaighofer, A., Lawrence, N.:
  Covariate shift and local learning by distribution matching (2008)

\bibitem{richter2016playing}
Richter, S.R., Vineet, V., Roth, S., Koltun, V.: Playing for data: Ground truth
  from computer games. In: European Conference on Computer Vision. pp.
  102--118. Springer (2016)

\bibitem{ros2016synthia}
Ros, G., Sellart, L., Materzynska, J., Vazquez, D., Lopez, A.M.: The synthia
  dataset: A large collection of synthetic images for semantic segmentation of
  urban scenes. In: Proceedings of the IEEE conference on computer vision and
  pattern recognition. pp. 3234--3243 (2016)

\bibitem{sankar2017unsuper}
Sankaranarayanan, S., Balaji, Y., Jain, A., Lim, S.N., Chellappa, R.:
  Unsupervised domain adaptation for semantic segmentation with gans. arXiv
  preprint arXiv:1711.06969  \textbf{2} (2017)

\bibitem{sankaranarayanan2018learning}
Sankaranarayanan, S., Balaji, Y., Jain, A., Nam~Lim, S., Chellappa, R.:
  Learning from synthetic data: Addressing domain shift for semantic
  segmentation. In: Proceedings of the IEEE Conference on Computer Vision and
  Pattern Recognition. pp. 3752--3761 (2018)

\bibitem{simonyan2014very}
Simonyan, K., Zisserman, A.: Very deep convolutional networks for large-scale
  image recognition. arXiv preprint arXiv:1409.1556  (2014)

\bibitem{tarvainen2017mean}
Tarvainen, A., Valpola, H.: Mean teachers are better role models:
  Weight-averaged consistency targets improve semi-supervised deep learning
  results. In: Advances in neural information processing systems. pp.
  1195--1204 (2017)

\bibitem{tsai2018learning}
Tsai, Y.H., Hung, W.C., Schulter, S., Sohn, K., Yang, M.H., Chandraker, M.:
  Learning to adapt structured output space for semantic segmentation. In:
  Proceedings of the IEEE Conference on Computer Vision and Pattern
  Recognition. pp. 7472--7481 (2018)

\bibitem{vu2018advent}
Vu, T.H., Jain, H., Bucher, M., Cord, M., P{\'e}rez, P.: Advent: Adversarial
  entropy minimization for domain adaptation in semantic segmentation. arXiv
  preprint arXiv:1811.12833  (2018)

\bibitem{wang2018understanding}
Wang, P., Chen, P., Yuan, Y., Liu, D., Huang, Z., Hou, X., Cottrell, G.:
  Understanding convolution for semantic segmentation. In: 2018 IEEE Winter
  Conference on Applications of Computer Vision (WACV). pp. 1451--1460. IEEE
  (2018)

\bibitem{yu2017dilated}
Yu, F., Koltun, V., Funkhouser, T.: Dilated residual networks. In: Proceedings
  of the IEEE conference on computer vision and pattern recognition. pp.
  472--480 (2017)

\bibitem{zhang2017curriculum}
Zhang, Y., David, P., Gong, B.: Curriculum domain adaptation for semantic
  segmentation of urban scenes. In: Proceedings of the IEEE International
  Conference on Computer Vision. pp. 2020--2030 (2017)

\bibitem{zhao2018icnet}
Zhao, H., Qi, X., Shen, X., Shi, J., Jia, J.: Icnet for real-time semantic
  segmentation on high-resolution images. In: Proceedings of the European
  Conference on Computer Vision (ECCV). pp. 405--420 (2018)

\bibitem{zhao2017pyramid}
Zhao, H., Shi, J., Qi, X., Wang, X., Jia, J.: Pyramid scene parsing network.
  In: Proceedings of the IEEE conference on computer vision and pattern
  recognition. pp. 2881--2890 (2017)

\bibitem{zhu2017unpaired}
Zhu, J.Y., Park, T., Isola, P., Efros, A.A.: Unpaired image-to-image
  translation using cycle-consistent adversarial networks. In: Proceedings of
  the IEEE International Conference on Computer Vision. pp. 2223--2232 (2017)

\bibitem{zou2018unsupervised}
Zou, Y., Yu, Z., Vijaya~Kumar, B., Wang, J.: Unsupervised domain adaptation for
  semantic segmentation via class-balanced self-training. In: Proceedings of
  the European Conference on Computer Vision (ECCV). pp. 289--305 (2018)

\end{thebibliography}
\end{document}